\documentclass[runningheads]{llncs}
\usepackage{graphicx}
\usepackage{float}
\usepackage{algorithmic}
\usepackage{amsmath, amsfonts, amssymb}
\usepackage[linesnumbered,ruled,vlined]{algorithm2e} 
\usepackage{pbox}
\usepackage{multirow}
\begin{document}
%
\title{Efficient Realistic Data Generation Framework leveraging Deep Learning-based Human Digitization}
%
%
\author{C. Symeonidis \and
P. Nousi \and
P. Tosidis \and
K. Tsampazis \and
N. Passalis \and
A. Tefas \and
N. Nikolaidis
 }

\authorrunning{C. Symeonidis et al.}
%
\institute{Artificial Intelligence and Information Analysis Lab, Department of Informatics, Aristotle University of Thessaloniki, Thessaloniki, Greece 
\email{\{charsyme,passalis,tefas,nnik\}@csd.auth.gr} }
\maketitle              
\begin{abstract}
The performance of supervised deep learning algorithms depends significantly on the scale,
quality and diversity of the data used for their training. Collecting and manually annotating large amount of data can be both time-consuming and costly tasks to perform. In the case of tasks related to visual human-centric perception, the collection and distribution of such data may also face restrictions due to legislation regarding privacy. In addition, the design and testing of complex systems, e.g., robots, which often employ deep learning-based perception models, may face severe difficulties as even state-of-the-art methods trained on real and large-scale datasets cannot always perform adequately due to not having been adapted to the visual differences between the virtual and the real world data. As an attempt to tackle and mitigate the effect of these issues, we present a method that  automatically generates realistic synthetic data with annotations for a) person detection, b) face recognition,  and c) human pose estimation. The proposed method takes as input real background images and populates them with human figures in various poses. Instead of using hand-made 3D human models, we propose the use of models generated through deep learning methods, further reducing the dataset creation costs, while maintaining a high level of realism. In addition, we provide open-source and easy to use tools that implement the proposed pipeline, allowing for generating highly-realistic synthetic datasets for a variety of tasks. A benchmarking and evaluation in the corresponding tasks shows that synthetic data can be effectively used as a supplement to real data.

\end{abstract}
\section{Introduction}
\label{section:Introduction}

 The scale, diversity and quality of data used for training supervised deep learning methods have a major impact on their performance. Algorithms that are intended to be deployed on real-life conditions are usually trained on multiple datasets in order to improve their  generalization abilities and ensure their robustness. 
 COCO \cite{COCODataset} and Cityscapes \cite{CITYSCAPESDataset} are only a few examples of large and diverse datasets providing annotations for training and evaluation of deep learning algorithms for computer vision tasks, such as object detection, human pose estimation, semantic image segmentation, etc. Collecting and manually annotating such amount of data is usually a challenging and exhausting task, requiring a lot of time and resources.  In the case of visual human-centric analysis, the collection and distribution of such data may also face restrictions due to legislation regarding privacy. An alternative approach for collecting training data in a more automated manner is to generate them through a simulator. Indeed, in recent years, the use of synthetic datasets generated in this way has been established on the computer vision domain~\cite{Gaidon2016CVPR}, \cite{Wang2019CVPR}, \cite{Tremblay2018CVPRWorkshops}. On the downside, these datasets often suffer in terms of realism and detail and/or are expensive to generate, requiring artists to carefully design specific models and environments.

Furthermore, apart from using synthetic data to create a new or enrich existing datasets, synthetic data are also often implicitly employed during the validation of complex systems, such as robots~\cite{michel2004cyberbotics}.  These systems are often designed and tested on simulation environments before being deployed in real life conditions. This allows for minimizing the risk of unwanted behaviors, that often lead to malfunctions that can destroy several hardware components, as well as for reducing development and validation time. However, deep learning methods trained on  datasets containing only real data often exhibit an unstable behaviour or fail to perform adequately due to not having been adapted to the visual differences between the simulated and the real world data, as we also experimentally demonstrate in this paper.

To overcome these limitations, in this paper we propose an effective and low-cost data generation method for human-centric tasks that:
\begin{itemize}    
    \item generates realistic and diverse data for person detection, pose estimation and face recognition, 
    \item is capable of reusing and augmenting existing datasets, eliminating additional costs that often occur, as well as potential data collection restrictions regarding privacy legislation, 
    \item provides automatically-generated and detailed annotations, and
    \item bridges the gap between the virtual and real world data, by enabling deep learning methods trained on the synthetic and real data to achieve highly accurate results on both.
\end{itemize}
At its core, the proposed pipeline for human-centric data generation uses as input real background images and carefully populates them with generated human figures in various poses. Instead of using hand-made 3D human models, we propose using models generated through deep learning methods. In this way, the proposed method eliminates most dataset creation costs, while maintaining a high level of realism. Finally, we provide an open-source implementation of the proposed method, allowing researchers and practitioners easily use the proposed method for generating highly-realistic synthetic datasets for a variety of tasks.

The rest of the paper is structured as follows. The related work is discussed in Section \ref{section:RelateWork}. We also provide a brief background on DL-based perception for the three tasks examined in this work in Section~\ref{section:background}. Then, the proposed method is described in detail in Section \ref{section:ProposedMethod}, while the dataset evaluation and benchmarking is provided in Section~\ref{section:DatasetEvaluation}. Finally,  conclusions are drawn in Section~\ref{section:Conclusion}.

\section{Related Work}
\label{section:RelateWork}

\subsection{Synthetic Data Generation for Computer Vision Methods}

The use of synthetic datasets has recently gathered pace in deep learning and such datasets have proven their value either as a replacement or as an augmentation to existing training data \cite{Alhaija2017BMVC,Richter2016PlayingFD}. Indeed, in \cite{Movshovitz2016ECCV}, it is demonstrated that deep neural networks can achieve state-of-the-art results when trained on synthetic, yet realistic datasets.  The domains of computer vision where synthetic datasets are widely used range from semantic image segmentation \cite{Ros2016CVPR}, and object detection \cite{Hinterstoisser2019ICCV} to pose estimation \cite{Tremblay2018CVPRWorkshops}, and face recognition~\cite{masi2017rapid}.
In \cite{Hattori2015CVPR}, the authors trained scene-specific pedestrian detection models using only synthetic data.  Surprisingly, those models managed to outperform models trained on real data. In \cite{Chen20163DV}, the authors presented an automatic approach that generates synthetic data for human pose estimation.  In \cite{Souza2017CVPR}, a diverse, large and realistic synthetic dataset for human action recognition was generated containing a total of 39,982 videos and 35 action categories. Finally, synthetic datasets are also used for facial recognition. Such an example is \cite{masi2017rapid}, where the authors use 3D models of faces to modify existing images in order to generate novel poses and expressions.

Furthermore, an alternative to using fully synthetic datasets is the generation of mixed reality datasets,  i.e.,  datasets  that  combine  synthetic  and  natural  content. In  \cite{Pishchulin2011CVPR},  the  authors  use 3D human models, which are rendered on random backgrounds in order to train a pedestrian detector.  In a similar fashion, the authors of \cite{Alhaija2018AugmentedRM} render 3D vehicle models into existing captured real world background images.  Inspired by these works, our method inserts realistic rendered 3D human models into existing natural background images, while trying to select appropriate scale and insertion locations, using a simple but effective approach. However, in contrast with the aforementioned approaches, the proposed method is capable of generating datasets for a wide variety of tasks, instead of being limited into a single tasks, e.g,. pedestrian detection~\cite{Pishchulin2011CVPR}.

\subsection{3D Human Model Generation from Natural Images}

Methods for 3D human model generation from images can be divided based on whether they use a single or multiple (-view) images as input. Most single view methods use parametric models \cite{Loper2015SMPLAS},  \cite{Pavlakos2019CVPR} of human bodies and shapes, due to the fundamental depth ambiguity. In \cite{Kanazawa2018CVPR}, the authors have shown that the pose and shape parameters of those models can accurately be estimated, so that each model fits and aligns perfectly with the figures of the depicted people. Usually, parametric models capture body characteristics and movements of naked human models. However, recent works using parametric models can also capture the shape \cite{alldieck2019tex2shape} and texture \cite{Lazova20193DV} of clothes, facial details, etc. Finally, the authors in \cite{saito2019pifu}, instead of using a parametric human model,  use an implicit representation that locally aligns pixels of 2D images with the global context of their corresponding 3D object. The method uses full-body images of people both from single and multiple views as input. 
In our work, we leverage a state-of-the-art 3D human generation model, i.e., the Pixel-aligned Implicit Function (PIFu) proposed in ~\cite{saito2019pifu}. This allows us to avoid most costs related to designing hand-crafted 3D human models. Instead, we dynamically create realistic 3D human models from existing datasets, that can be subsequently used for generating realistic and diverse datasets.

\subsection{Deep Learning for Human-centric Perception}
\label{section:background}

Person detection constitutes a crucial task related to human-centric perception, where the models aim to detect whether one or more persons exist in a given image, as well as to localize them. Given the proper training annotations, which are usually supplied in the form of bounding box coordinates for each person that appears in an image, generic object detection methods can be trained, such as  the Single Shot Detector (SSD) \cite{liu2016ssd}. It is also worth noting that person detection has been studied separately in recent literature, due in part to the single-class nature of this task. This can also for the development of specialized methods, which can benefit from prior knowledge surrounding the task~\cite{rodriguez2011density}.

Face recognition refers to analyzing images where humans are depicted in order to extract their identities. 
Three modules are usually needed for a complete face recognition system: a) a face detector to localize faces in a given image, b) a face alignment method, to ensure the proper alignment of the cropped facial image, and c) a feature extractor that is used to extract discriminative features from each detected face. These features are then fed to a matching algorithm to determine the identity of the face, by computing similarity scores against a database, or used by a classifier to directly predict the identity of persons. Most of the recent literature focuses on training deep learning models  by using metric learning objectives, such as angular/cosine-margin-base losses, e.g, 
CosFace~\cite{wang2018cosface}, and  ArcFace~\cite{deng2019arcface}. 
These methods aim to maximize inter-class variance and minimize intra-class variance in the resulting representation space. The annotations used for training face recognition approaches are similar to any classification task, i.e., each face appearing in an image must be annotated by its name (or a pseudo-anonymized id). Typically, these annotations are then used for evaluation by forming pairs of images that  correspond either  to the same person or to different persons. 
The most commonly used architectures in recent face recognition systems employ deep models, such as Residual Neural Networks (ResNets)~\cite{he2015deep} and MobileNets (MobileFaceNet)~\cite{chen2018mobilefacenets}

Human pose estimation refers to  analyzing images or videos of humans to predict their pose, which in essence entails the localization of their joints, allowing for inferring skeleton information. Pose estimation is a key step for enabling machines to understand peoples' actions and intentions in various robotics applications~\cite{zimmermann20183d}. 
To the best of our knowledge, among the most accurate method currently is the Distribution-Aware coordinate Representation of Keypoint (DARK) method \cite{zhang2019distributionaware}, which serves as a plugin for other deep learning-based pose estimation methods, improving the results by modifying the coordinate representation of the heatmap and also by changing the way the ground-truth coordinates are converted to heatmaps. Furthermore, for single pose estimation the soft-gated method presented in~\cite{bulat2020fast}, which introduces gated skip connections with per-channel learnable parameters to control the data flow for each channel, achieved state-of-the-art results This method also employs a complex hybrid network that combines the HourGlass and U-Net architectures to further improve the accuracy of pose estimation. 


For all the deep learning architectures used by person detection, face recognition and pose estimation large-scale datasets are employed during the training process. This is required in order to ensure that the models will  be trained effectively, avoiding over-fitting phenomena. The emergence of large-scale datasets, such as COCO~\cite{COCODataset} for person detection and pose estimation and Microsoft Celeb dataset (MSCeleb)~\cite{msceleb} for face recognition, enabled training these models by overcoming these issues. However, significant concerns have been raised recently, mainly regarding maintaining the privacy of the human subjects that appear in these datasets~\cite{girasa2020ethics}. As a results, creating new datasets for these tasks is costly and requires a significant effort, while, in some cases, even the use of existing datasets might be limited due to ethics concerns.


\section{Proposed method}
\label{section:ProposedMethod}

The proposed method for realistic synthetic data generation for human-centric tasks consists of two stages. First, 3D human models are generated and  skeleton-related information is extracted for each of them. For this task, we employ natural images, avoiding the need for using hand-crafted 3D models. Then, the proposed pipeline proceeds with the second stage, which concerns data generation through careful blending of real background images and 3D human models. The appropriate constraints are enforced during this step to ensure realistic placement of objects. In the remaining of this Section, we analytically describe each of these two steps.

\subsection{3D Human Model Generation and Skeleton Extraction }

\begin{figure}[t!]
    \centering
    \centerline{
    \includegraphics[width=0.92\textwidth]{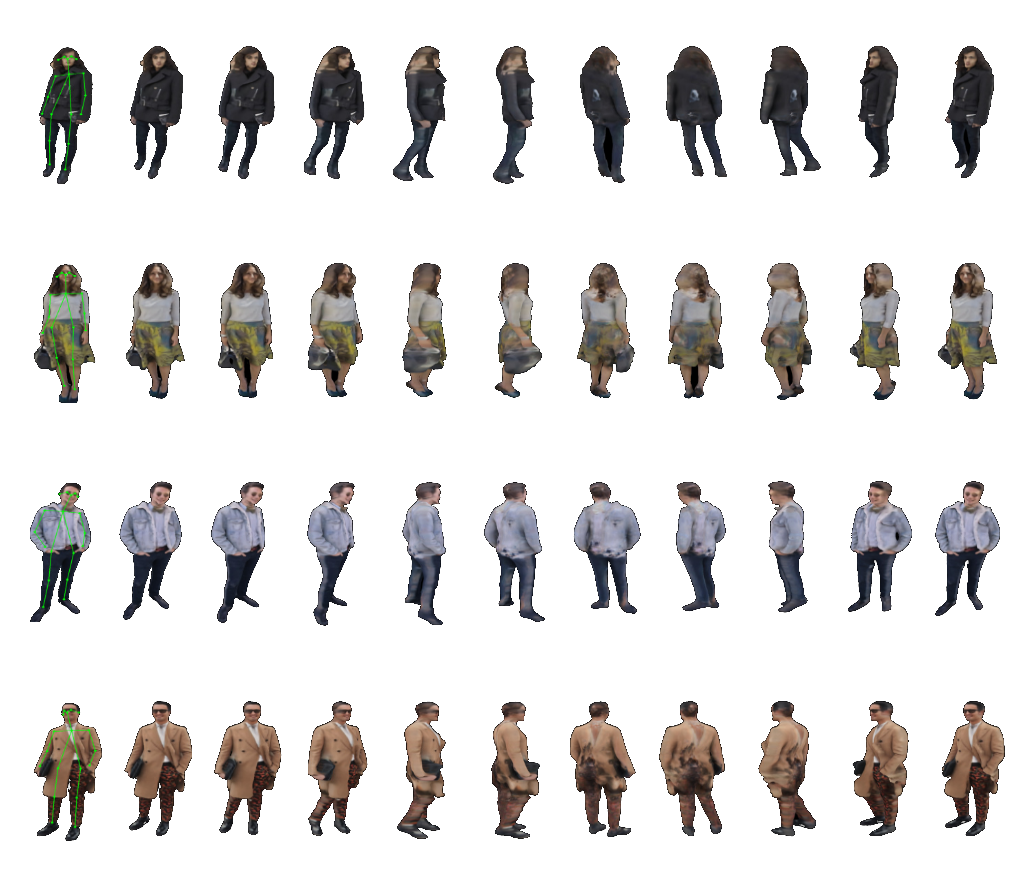}
    }
    \caption{3D human models in various views generated by PIFu.  Full-body images from the CCP dataset were used as input. }
    \label{fig:human_figures}
\end{figure}

After a careful review of the relevant literature, PIFu \cite{saito2019pifu} was selected as the most suitable deep learning method for realistic 3D human model generation from single-view images.
Although the 3D human models generated from PIFu, achieve a high level of detail they are not articulated, which means that they cannot be directly used for pose estimation tasks. To overcome this limitation, we propose approximating the 3D positions of the joints by employing multiple renderings of the human models from multiple views. Then, an off-the-shelf pose estimator, the OpenPose estimator  \cite{osokin2018lightweight_openpose}, is employed to estimate the 3D positions of the joints. Increasing the number of views used for estimating the 3D positions of the joints can significantly increase the accuracy of the generated annotations. The corresponding algorithmic procedure is described on  Algorithm \ref{alg:3dpose_est}. Some typical examples of images of human models that were generated using the proposed method are depicted in Figure \ref{fig:human_figures}.

 \begin{algorithm}
 \small{\caption{3D pose estimation through 2D pose estimation using multi-view images}}
     \label{alg:3dpose_est}
     \textbf{Input}: Initial camera poses $\mathbf{D}=[\mathbf{d}_1,...,\mathbf{d}_N]^T \in \mathbb{R}^{N \times 6}$  \\
    \textbf{Output}: Estimated 3D joint positions $\mathbf{J}^{3D}=[\mathbf{j}^{3D}_1,..,\mathbf{j}^{3D}_S]^T \in \mathbb{R}^{S \times 3} $  \\
    \For {$\mathbf{d}_i \in \mathbf{D}$}{
        Set the camera in pose $\mathbf{d}_i$ \\
        $\mathbf{I}_i \leftarrow $ \textup{Render the image}\\
        $\mathbf{J}^{2D}_i \leftarrow $ \textup{Perform pose estimation on image } $\mathbf{I}_i$ \\
        $[\mathbf{V}^{near},\mathbf{V}^{far}]^T \leftarrow $  \textup{Unproject $\mathbf{J}^{2D}_i$ in far and near planes } \\
        $\mathbf{L}^{3D}_i \leftarrow $ \textup{For each joint compute a line in 3D space given $[\mathbf{V}^{near}, \mathbf{V}^{far}]^T$ }
    }
    \For {$s \gets 1 $  to  $S$} {
        $\mathbf{j}^{3D}_s \leftarrow $ Compute the position of a joint in 3D space using Least-Squares approximation on lines $\mathbf{L}^{3D}$
    }
    \Return{$\mathbf{J}^{3D}$}
    
 \end{algorithm}

\subsection{Data  generation  through  synthesis  of  real  background  images  and 3D human models}

\begin{figure}
    \centering
    \centerline{\includegraphics[width=1.0\textwidth]{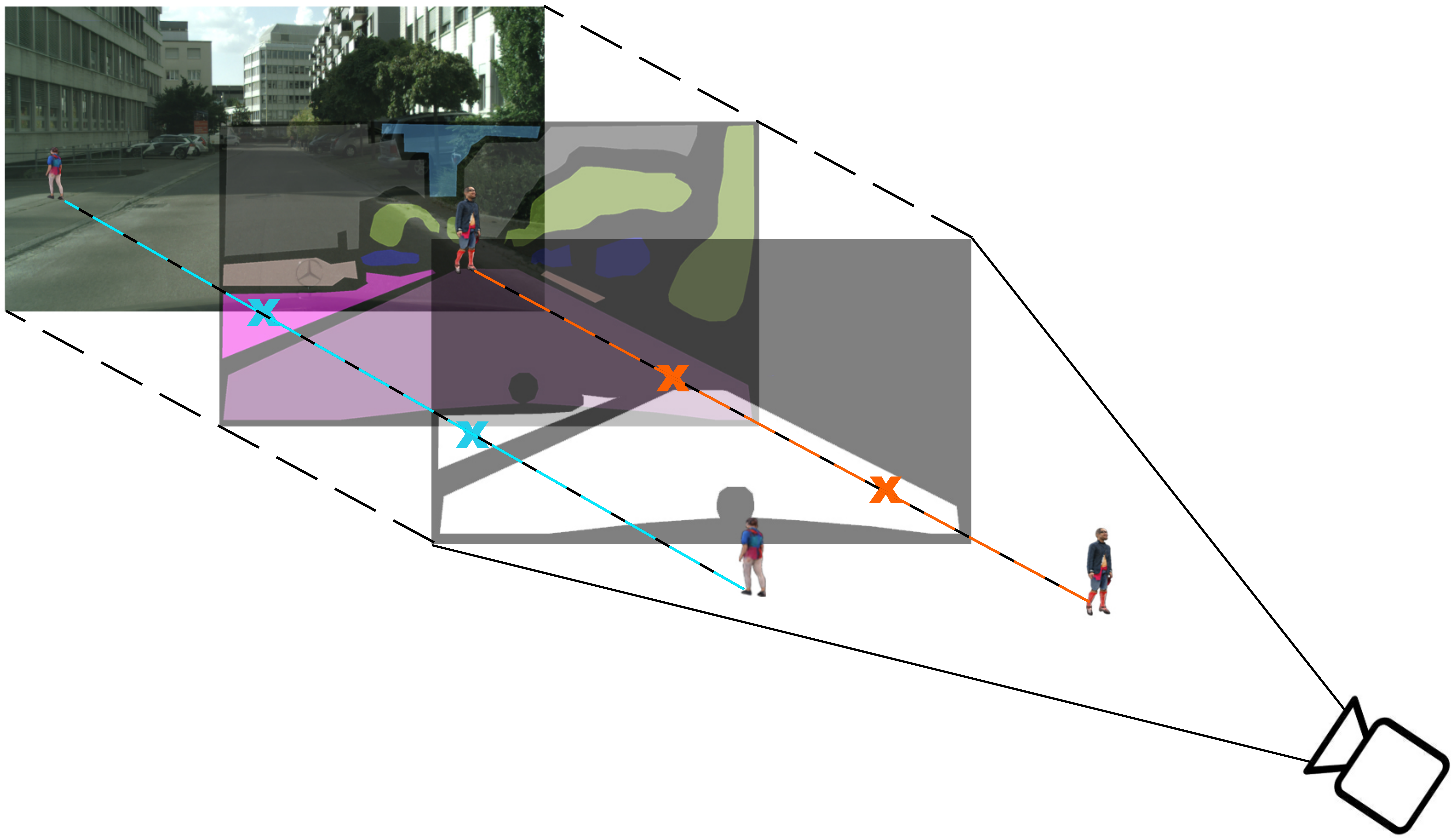}}
    \caption{Visualization of the data generation process.}
    \label{fig:proposed}
    \vspace{-0.5em}
\end{figure}

The proposed data generation method is implemented in a virtual 3D environment using OpenGL. It receives as input a set of natural background images and a set of 3D human models. Semantic image segmentation ground truth information must also be provided for the corresponding background images. If this information is not available, then the predictions from a semantic image segmentation method, such as \cite{Zhao_2017_CVPR}, can be used instead. As shown in Fig.~\ref{fig:proposed}, the proposed method works by projecting different human 3D model into different, yet valid, places of a background image. More specifically, the virtual environment  consist of an image projection plane and a virtual camera that is placed and properly calibrated so as the background images are fully visible when projected to the plane. 

During the process of data generation, the method selects a background image and projects it to the plane. Then, potential 2D image locations for human figure placement are computed, based on the semantic image segmentation ground truth. For example, in the case of an outdoor environment, pixels that correspond to classes such as ``road'' or ``pavement'', can be selected for human figure placement. Afterwards, randomly selected 3D human models are placed in various poses between the plane and the camera in a way that their projection coincides with the potential 2D image locations for human figure placement that were computed in the previous steps. The projected scale of the models is modified, by controlling the distance between the camera and the human figure, considering the rules of perspective projection, so that the images captured from the virtual camera look natural and realistic. Finally, the ground truth annotations required for pose estimation methods is computed through the projection of the 3D locations of the joints of the populated models. Other annotation types, e.g., bounding boxes, can be trivially computed in a similar way.

\section{Experimental Evaluation}
\label{section:DatasetEvaluation}

\subsection{Dataset Generation}
The proposed dataset generation pipeline was evaluated by generating a large-scale synthetic dataset. To this end, the 3D human models were generated using a pre-trained model of PIFu and full-body images of people from the Clothing Co-Parsing (CCP) \cite{yang2014clothing} dataset. In particular,  we were able to generate 133 human models, each one corresponding to a different human. As background images for our environment we selected images from Cityscapes dataset~\cite{CITYSCAPESDataset}. The Cityscapes dataset contains a diverse set of video sequences recorded in street scenes from 50 different cities. We selected to use a subset of 20,000 images, for which coarse annotations for semantic image segmentation were provided. The number of the images that were actually used was smaller, as images already containing humans were dropped. Overall, the generated dataset contains 50,000 images. The training set contains 40,000 images, while the rest form the test set.  The generated data were used for three different tasks, i.e., person detection, face recognition and human pose estimation and evaluated using a wide range of different settings and setups.



\subsection{Experimental Evaluation} 

First, we report results on person detection using two different object detection methods and backbones, all pretrained on the COCO dataset~\cite{COCODataset}.  This first set of experiments aims to evaluate the ability of pretrained models to effectively detect persons that appear in the generated data, examining in this way whether the generated data are realistic enough. The results are reported in Table~\ref{table:persondetection}, in terms of precision at 0.5 IoU, for the person class of the COCO validation set as well as for the test set of the generated dataset. All detectors exhibit excellent precision on person detection tasks, validating the aforementioned hypothesis. It should be noted that, apart from CenterNet, the detectors generalize better on this dataset compared to the validation set of COCO.

\begin{table}[t!]

\begin{minipage}[b]{.5\textwidth}
\caption{Person Detection Evaluation (AP@0.5 is reported)}
\small
\label{table:persondetection}
\begin{center}
\begin{tabular}{l|c|c}
 \textbf{Model} & \textbf{\ COCO \ } & \textbf{\ Synthetic}\\ \hline
CenterNet (RS-18) & 38.2\% & 39.0\% \\ \hline 
SSD-512  (RS-50) & 41.9\% & 86.8\% \\ \hline
SSD-512 (VGG16) & 41.5\% & 88.1\% \\ 
\end{tabular}
\scriptsize{``RS'' refers to ResNet.}
\end{center}
\hspace{0.1em}
\end{minipage}
\begin{minipage}[b]{.48\textwidth}
\caption{Human Pose Estimation Evaluation (AP@0.5 is reported)}
\label{table:pose}
\begin{center}
\footnotesize
\begin{tabular}{l|c|c}
 \textbf{Data \ } & \textbf{\ COCO\ } & \textbf{\ Synthetic\ }\\ \hline
R & 37.7\% & 52.7\% \\ \hline 
S & 1.2\% & 93.3\% \\ \hline
R+S  & 34.0\% & 88.7\% \\ 
\end{tabular}
\end{center}
\scriptsize{
``R'' refers to using real data for training (COCO dataset), ``S'' refers to using the proposed synthetic data and ``R+S'' to using both of them.}
\end{minipage}
\end{table}

Similar results are also reported for pose estimation using the lightweight OpenPose model~\cite{osokin2018lightweight_openpose}. The results are reported in Table~\ref{table:pose}. For these experiments, instead of evaluating the precision of different pretrained models, we evaluate the effect of using only real data for training (``R''), using only synthetic data (``S'') and combining both of them for the training process (``R+S'').  The training process ran for 140,000 iterations for all models following the setup proposed in~\cite{osokin2018lightweight_openpose}.
First, note that the ability of pretrained pose estimators to generalize on the data is validated (``R'' row). Then, training only with synthetic data leads to perfect generalization on the test set of the synthetic data, but fails to generalize on real data. On the other hand, when both synthetic and real data are combined, we observe a significant improvement on the precision on the synthetic test set. A small decrease is observed on the COCO test, but this is probably due to using a lightweight model architecture with relatively low learning capacity. 

Finally, we evaluated the generated data for the task of face recognition, where we observed several interesting phenomena. The results are reported in Table~\ref{table:face}. The used models were a) pretrained on the MSCeleb (trained for 120 epochs), b) trained both on the combined MSCeleb and the proposed dataset (trained for 120 epochs), c) pretrained on the MSCeleb data and fine-tuned on the combined dataset (trained for 5 epochs). An inverted residual model (IR-50) was used as feature extractor~\cite{behrmann2019invertible}. The evaluation datasets consisted of: a) 5,000 positive and 5,000 negative pairs of images taken from the MSCeleb dataset, b) 5,000 positive and 5,000 negative pairs of images of the test set of the generated synthetic data, c) the LFW dataset~\cite{LFWTech} and d) the CFP\_FF dataset~\cite{cfp-paper}.

Several interesting conclusions can be drawn from the results reported in Table~\ref{table:face}. First, note that the synthetic data are actually ``harder'' for a network trained on real data (``R'' row). This can be explained if we consider the loss of facial detail that often occur in the 3D models.  Quite interesting, this is also the case for real face recognition systems that operate \textit{in-the-wild} with low resolution cameras (e.g., footage from CCTV systems), as well as for robotics systems that are validated using simulators. Then, we observe that training using both the combined real and synthetic set leads to tremendous improvements both for the real and synthetic data, without any significant impact on how the model generalizes on other real datasets (e.g., LFW and CFP\_FF). Finally, we observed similar positive results even when we only fine-tuned the model trained for 5 epochs using the combined set (last row), demonstrating that we can obtain similar improvements with full training (``R+S'') only spending a fraction of time for training (5 epochs instead of 120).

\begin{table}[t!]
\caption {Face Recognition Evaluation (face verification precision (\%) is reported)}
\vspace{-1.5em}
\label{table:face}
\begin{center}
\begin{tabular}{c|cccccccc}
 {\textbf{Training Setup}} & { \ \ \textbf{MSCeleb} \ \ } & {\ \ \textbf{Synthetic Data}\ \ } & {\ \ \textbf{LFW}\ \ } & {\ \ \textbf{CFP\_FF} \ \ } \\  \hline
{R} & 95.01\%  &  90.63\% & 99.80\% & 99.67\%  \\ \hline
{R+S}& 98.44\% & 99.75\% &  99.20\%  & 99.05\% \\ \hline
{R+S (finetuning)}& 97.70\%  & 99.47\% &  99.38\% &  99.27\% \\ \hline
\end{tabular}
\end{center}
\end{table}

\section{Conclusion}
\label{section:Conclusion}

In this paper we presented a method capable of automatically generating realistic synthetic data with annotations, while leveraging existing images both for the creation of 3D models of humans, as well as for employing realistic backgrounds. The proposed method has minimal cost, since it does not require handcrafted 3D models or simulation environments and it can be used for generating data for a variety of tasks, such as  person detection, face recognition, and human pose estimation. The conducted experimental evaluation demonstrated that the generated data are suitable for training DL models in most cases. It is worth noting that for some tasks, such as face recognition, further improvements can be obtained. These results highlight the potential of the proposed data generation method both for generating large-scale datasets for data-scarce domains, as well as for minimizing the distribution shift that is experienced in many robotics applications, e.g., when transferring the models from a simulation environment in real deployment and vice versa.
indent 
\\
\noindent \textbf{Acknowledgment} This project has received funding from the European Union's Horizon  2020  research  and  innovation  programme  under grant agreement No 871449 (OpenDR). This publication reflects the authors views only.  The European Commission is not responsible for any use that may be made of the information it contains

This is a preprint of the following chapter: C. Symeonidis, P. Nousi, P. Tosidis, K. Tsampazis, N. Passalis, A. Tefas and N. Nikolaidis, Efficient Realistic Data Generation Framework Leveraging Deep Learning-Based Human Digitization, published in Proceedings of the 22nd Engineering Applications of Neural Networks Conference, edited by L. Iliadis, J. MacIntyre, C. Jayne and E. Pimenidis, 2021, Springer, Cham reproduced with permission of Springer Nature Switzerland AG 2021. The final authenticated version is available online at: https://doi.org/10.1007/978-3-030-80568-5.

\bibliographystyle{splncs04} 
\bibliography{bib}

\end{document}